\documentclass{article}

\usepackage{microtype}
\usepackage{graphicx}
\usepackage{subcaption}
\usepackage{booktabs} % for professional tables
\usepackage{xcolor}         % color comments
\usepackage{enumerate}
\usepackage{amsmath, amsthm, amssymb}
\usepackage{natbib}

\newif\ifsubmit
\submittrue
\ifsubmit
\newcommand{\dnote}[1]{}
\newcommand{\mnote}[1]{}

\else
\newcommand{\dnote}[1]{\textcolor{olive}{\textbf{Dilip: #1}}}
\newcommand{\mnote}[1]{\textcolor{green}{\textbf{Michael: #1}}}

\fi

\usepackage{hyperref}

\usepackage[accepted]{icml2018}

\icmltitlerunning{Deep Reinforcement Learning from Policy-Dependent Human Feedback}
\begin{document}
% \maketitle
\twocolumn[
\icmltitle{Deep Reinforcement Learning from Policy-Dependent Human Feedback}
% \title{Deep Reinforcement Learning from Policy-Dependent Human Feedback}




% \begin{icmlauthorlist}
% \icmlauthor{Dilip Arumugam\footnote}{st}
% \icmlauthor{Jun Ki Lee\footnotemark}{br}
% \icmlauthor{Sophie Saskin\footnotemark[\value{footnote}]}{br}
% \icmlauthor{Michael L. Littman\footnotemark[\value{footnote}]}{br}
% \end{icmlauthorlist}

\begin{icmlauthorlist}
\icmlauthor{Dilip Arumugam}{st}
\icmlauthor{Jun Ki Lee}{br}
\icmlauthor{Sophie Saskin}{br}
\icmlauthor{Michael L. Littman}{br}
\end{icmlauthorlist}

\icmlaffiliation{st}{Department of Computer Science, Stanford University}
\icmlaffiliation{br}{Department of Computer Science, Brown University}

\icmlcorrespondingauthor{Dilip Arumugam}{dilip@cs.stanford.edu}
% \icmlcorrespondingauthor{Michael L Littman}{mlittman@cs.brown.edu}

% \icmlkeywords{Reinforcement learning; Deep learning; Human-robot/agent interaction; Reward structures of learning}

\vskip 0.3in
]



\printAffiliationsAndNotice{}  % leave blank if no need to mention equal contribution
% \maketitle

\begin{abstract}
% \footnotetext{Department of Computer Science, Stanford University.}
% \footnotetext{Department of Computer Science, Brown University.}

To widen their accessibility and increase their utility, intelligent agents must be able to learn complex behaviors as specified by (non-expert) human users. Moreover, they will need to learn these behaviors within a reasonable amount of time while efficiently leveraging the sparse feedback a human trainer is capable of providing. Recent work has shown that human feedback can be characterized as a critique of an agent's current behavior rather than as an alternative reward signal to be maximized, culminating in the COnvergent Actor-Critic by Humans (COACH) algorithm for making direct policy updates based on human feedback. Our work builds on COACH, moving to a setting where the agent's policy is represented by a deep neural network. We employ a series of modifications on top of the original COACH algorithm that are critical for successfully learning behaviors from high-dimensional observations, while also satisfying the constraint of obtaining reduced sample complexity. We demonstrate the effectiveness of our Deep COACH algorithm in the rich 3D world of Minecraft with an agent that learns to complete tasks by mapping from raw pixels to actions using only real-time human feedback in 10--15 minutes of interaction.
\end{abstract}

\section{Introduction}
\label{sec:intro}

Recent years have seen breakthrough successes in the area of reinforcement learning, leveraging deep neural networks to learn complex behaviors from high-dimensional data \cite{Mnih2015HumanlevelCT,Lillicrap2015ContinuousCW,Silver2016MasteringTG,Schulman2017ProximalPO,Hessel2018RainbowCI}. These approaches leverage a hard-coded reward function that encapsulates the underlying task to be solved. While this setting is suitable for a wide range of applications, certain behaviors are not so easily expressed or derived as reward functions \cite{Littman2017EnvironmentIndependentTS}. Furthermore, as these intelligent decision-making agents become more ubiquitous and find themselves placed in novel environments alongside non-expert users, the need will arise for a simplified end-user interaction that does not require low-level programming knowledge or careful incentive engineering.

\begin{figure}
\centering
\includegraphics[width=\linewidth]{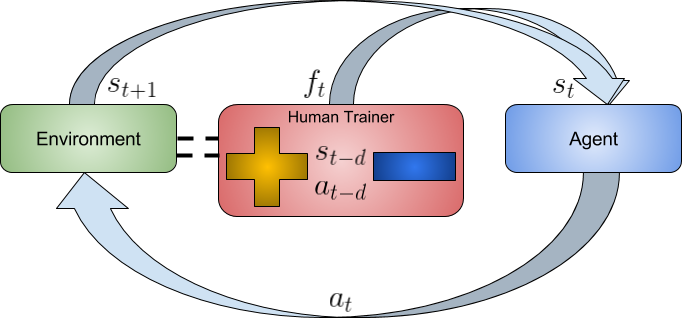}
\caption{Depiction of the human-in-the-loop reinforcement-learning setting. Instead of a reward signal from the environment, the agent receives a feedback signal, $f_t$, from a human trainer based on observing the agent's behavior in the environment. There is a natural latency in the human response time such that the human is always evaluating the behavior from $d$ timesteps ago.}
\label{fig:rl}
\end{figure}

To address these concerns, we turn to the area of \textit{human-in-the-loop reinforcement learning} (HRL) \cite{Amershi2014PowerTT}, which mimics the traditional reinforcement-learning setting in all regards except for the specification of learner feedback; in lieu of a hard-coded reward function, HRL algorithms respond to positive and negative feedback signals as provided by a human trainer who is ``in the loop'' during the learning process. Beyond allowing for the provision of more informative feedback than a traditional, hard-coded reward function, this setting also bypasses the need for end-users to understand the inner workings of the agent or how to write computer programs.

While there have been varied approaches to tackling HRL problems~\cite{Thomaz2006ReinforcementLW,Knox2008TAMERTA,Griffith2013PolicySI,Loftin2015LearningBV,Knox2010CombiningMF,Vien2012ReinforcementLC}, we extend an approach that is exceptionally considerate of how human trainers provide feedback to learning agents. Built upon the realization that human feedback can be treated as an unbiased estimate of the advantage function, COnvergent Actor Critic by Humans (COACH)~\cite{MacGlashan2017InteractiveLF} is a simple actor-critic reinforcement-learning algorithm that supports learning directly from human feedback. In this work, we study the efficacy of COACH when scaling to more complex domains where higher dimensional data demands the use of nonlinear function-approximation techniques for success. 
% The transition to deep neural network-based policies 
This transition requires a series of additions on top of the base COACH algorithm so as to maintain low sample complexity (in terms of human feedback signals) while supporting robust learning. We conduct an evaluation of our extension, Deep COACH, against the original COACH and the related Deep TAMER \cite{Warnell2017DeepTAMER} algorithms with experiments run in the 3D game world of Minecraft where all behaviors are to be learned directly from raw pixel observations and human feedback. 
% To the best of the authors' knowledge, we are the first to demonstrate a HRL approach that learns multiple behaviors within a single domain.

\section{Related Work}
\label{sec:relatedwork}
Our approach is directly inspired by the COACH algorithm~\cite{MacGlashan2017InteractiveLF}, which recognizes that a human trainer observing an agent's execution provides feedback contingent on the demonstrated behavior. Furthermore, through a set of user studies, the authors illustrate that human feedback exhibits properties that are inconsistent with a traditional reward signal. For instance, a reward function will always provide an agent with positive feedback for demonstrating correct behavior whereas real human feedback exhibits a diminishing returns property; as the agent settles into correct behavior, the human trainer becomes increasingly less likely to provide redundant positive feedback for repetitive good behavior. 
%This along with other properties suggests that human feedback can be approximated by the advantage function, $A^{\pi}(s,a)$. 
This, along with other properties, suggests that human feedback is more accurately depicted as an evaluation of an agent's action choice in the context of its current behavior. With this realization, the interaction between learning agent and human trainer can be represented within an actor-critic reinforcement-learning algorithm where the human trainer is the critic evaluating the actor's current policy. While the authors presented a generic, real-time COACH algorithm, empirical evaluations only used hand-coded image feature detectors for observations to only learn policies via linear function approximation. Our approach successfully scales up to problem settings where observations are high-dimensional and more powerful function-approximation techniques are required to express the desired behavior.

The TAMER framework~\cite{Knox2008TAMERTA} provides an alternate approach to reinforcement learning from human feedback by training a regression model to represent a reward function consistent with the feedback signals provided by a human trainer. Under this paradigm, the learner is not as limited by the capacity of the human trainer to provide feedback so long as a sufficient number of samples is collected for learning an accurate regression model. Despite these advantages, the TAMER framework still fails to account for the nature of human trainers or the structure of human feedback, which can result in the forgetting of learned behavior~\cite{MacGlashan2017InteractiveLF}. Recently, \citeauthor{Warnell2017DeepTAMER}~(\citeyear{Warnell2017DeepTAMER})  proposed Deep TAMER for neural-network-based HRL that builds on top of the TAMER framework. While there are some similarities between Deep TAMER and Deep COACH, the base approaches of TAMER and COACH take fundamentally different stances on the HRL problem. Furthermore, it is initially unclear how sample efficient the proposed Deep TAMER approach is; while being able to achieve satisfactory performance on a chosen arcade game task, there are no results reporting the total number of samples needed by human trainers to elicit the desired behavior. In contrast, we include figures for all experiments illustrating the breakdown of trainer feedback for both Deep COACH and Deep TAMER in terms of both positive and negative signals over a set of independent trials.
% As training progresses, 

\citeauthor{Christiano2017DeepRL}~(\citeyear{Christiano2017DeepRL}) also proposed a TAMER-inspired approach for learning from human preferences; specifically, they asynchronously learn a reward model through data collected from a human observing and noting preferences between short agent trajectories. After these human preferences are translated into scalar reward signals and used as training data for a regression model, what remains is a traditional reinforcement-learning problem. Most notably, our work differs in that we apply deep reinforcement learning based on direct human feedback instead of attempting to learn human preferences. Furthermore, even with only hundreds or thousands of samples collected from real humans (or from a synthetic oracle), the approach presented by \citeauthor{Christiano2017DeepRL} still requires millions of agent steps until traditional reinforcement-learning algorithms converge on a satisfactory policy for certain tasks. While we opt to not perform experiments in Atari domains, we suspect that the lack of a dependence on traditional deep reinforcement-learning algorithms offers room for greater sample efficiency. 
% Certainly, the experiments in this work demonstrate that Deep COACH can learn complex behaviors from high dimensional observations using less than one hundred human feedback samples and within hundreds of agent steps in the world.

Other related work of interest includes that of \citet{Griffith2013PolicySI} which, much like \citet{Loftin2015LearningBV}, takes a third perspective on HRL that views human feedback as a label of action optimality. Moreover, the approaches of policy shaping \citep{Griffith2013PolicySI} and SABL \citep{Loftin2015LearningBV} integrate information about the human trainer based on observed feedback to improve learning. Other variants of TAMER have since expanded to integrate simultaneous learning from both human reward signals in addition to a base reward function for the underlying MDP \citep{Knox2010CombiningMF,Knox2012ReinforcementLF} along with modifications to handle continuous state-action spaces \citep{Vien2012ReinforcementLC}. In contrast to these approaches, (Deep) COACH views feedback as a relative judgement of recent behavior in comparison to the trainer's desired policy. 

Lastly, we make a distinction between the HRL setting utilized in this work and the complementary area of learning from demonstration~\cite{Argall2009ASO}. Under the learning-from-demonstration paradigm, an agent is provided with a dataset of demonstrations (usually, as trajectories), which capture a desired behavior. Given such a dataset, a learner is meant to solve for the underlying policy (or associated reward function) that best aligns with the observed data. Great successes in this area include imitation learning~\cite{Ross2011ARO}, inverse reinforcement learning~\cite{Abbeel2004ApprenticeshipLV}, and inverse optimal control~\cite{Finn2016GuidedCL}. In scenarios where it is possible to obtain such valuable demonstration data, these kinds of learning-from-demonstration approaches could be substituted for the unsupervised pre-training phase of our Deep COACH algorithm, leading to a robust initial policy that a human trainer could gradually augment with live feedback as needed.

Previous work that has demonstrated the utility of Minecraft as a testbed for machine-learning research~\cite{Aluru2015MinecraftAA,Tessler2016ADH,Oh2016ControlOM}.
%We assert that one of the core goals of human-in-the-loop reinforcement learning is to allow a non-expert trainer to guide the behavior of an agent in a rich domain where many possible behaviors exist. 
Unlike more traditional domains built around arcade video games with single objectives, Minecraft presents a richer environment with more complicated dynamics and a consistent, voxelized representation that more closely emulates the complexity of the real world. To leverage these features, all of our experiment domains were developed within the Project Malm{\"o} platform\footnote{https://github.com/Microsoft/malmo} \cite{Johnson2016TheMP}, which allows for the creation and deployment of AI experiments within Minecraft. 

\section{Approach}

We begin with a summary of background information before dissecting the components of our Deep COACH algorithm.

\label{sec:approach}
\subsection{Background}
\label{sec:background}
Within the HRL setting, we turn to the Markov Decision Process (MDP)~\cite{Puterman94} formalism for representing the underlying sequential decision-making problem. Specifically, an MDP is denoted by the $\langle \mathcal{S}, \mathcal{A}, \mathcal{T}, \mathcal{R}, \gamma \rangle$ five-tuple, where $\mathcal{S}$ denotes a set of states, $\mathcal{A}$ denotes the set of actions available to the agent, $\mathcal{T}$ specifies the transition probability distribution over all states given the current state and selected action, $\mathcal{R}$ denotes the reward function, and $\gamma$ is the discount factor. The goal of a learning agent is to select actions at each timestep $t$ according to a policy, $\pi$, so as to maximize total expected discounted reward. Here, we take $\pi$ to be a stochastic policy parameterized by $\theta_t$, denoted $\pi_{\theta_t}$, which defines a probability distribution over all actions given the current state. Note that, within the HRL paradigm, the agent has no access to the true environment reward signals as specified by $\mathcal{R}$ (even if it exists and can be written down) and is instead only given access to the feedback signal of a human trainer at each timestep, $f_t \in \{-1, 0, 1\}$,  where $f_t=0$ represents no provided feedback. 

MDPs naturally give rise to two important functions.
% , the state--value function and the action--value function.
The state--value function, $V^{\pi}(s)$, intuitively captures the utility of a particular state as measured by the expected discounted sum of future rewards to be earned from that state by following the current policy, $\pi$. Similarly, the action--value function, $Q^{\pi}(s,a)$, captures the quality of a particular deviation from the current policy in a given state as measured by the expected discounted sum of future rewards to be earned from taking the specified action and then following the policy $\pi$ thereafter. Formally, both $V^{\pi}(s)$ and $Q^{\pi}(s,a)$ are defined by the Bellman equations where $V^{\pi}(s) = \mathbb{E}_{a \sim \pi}[Q^{\pi}(s,a)]$ and $Q^{\pi}(s,a) = \mathcal{R}(s,a) + \gamma\mathbb{E}_{s^{'} \sim \mathcal{T}(\cdot|s,a)}[V^{\pi}(s^{'})]$.

One widely used class of algorithms for policy-based reinforcement learning are \textit{actor-critic} \cite{Sutton1999PolicyGM} algorithms, where learning occurs from the interplay between two distinct models, an actor and a critic. The actor is a parameterized policy, $\pi_{\theta_t}$, used by an agent for action selection, whereas the critic is a parameterized value function, $Q^{\pi_{\theta_t}}(s,a)$, induced by the actor's policy. As the critic evaluates the policy being executed by the agent, it informs the actor of the direction in which to move the policy parameters so as to improve performance and maximize returns. The policy achieves a discounted sum of rewards, $R_t = \sum\limits_{t=1}^\infty \gamma^{t-1} r_t$, that is needs to optimize using the objective $J(\theta_t) = \mathbb{E}_{\pi_{\theta_t}}[R_t]$. \citeauthor{Sutton1999PolicyGM}~(\citeyear{Sutton1999PolicyGM}) prove that this objective induces the following policy gradient:
\begin{align}
\label{eq:pg}
\nabla_{\theta_t} J(\theta_t) &= \mathbb{E}_{a \sim \pi_{\theta_t}(\cdot | s)}[\nabla_{\theta_t} \log \pi_{\theta_t}(a | s) Q^{\pi_{\theta_t}}(s,a)].
\end{align}
While the information offered by the critic is adequately captured by the action--value function, \citeauthor{Sutton1999PolicyGM} note that the subtraction of an action-independent baseline function leads to an equivalent formulation. Consequently, we can use the state--value function $V^{\pi_{\theta_t}}(s)$ as a baseline function and reformulate the objective in Equation~\ref{eq:pg} as:
% \begin{align}
% \nabla_{\theta_t} J(\theta_t) &= \mathbb{E}_{a \sim \pi_{\theta_t}(\cdot | s)}[\nabla_{\theta_t} \log \pi_{\theta_t}(a | s) (Q^{\pi_{\theta_t}}(s,a) - V^{\pi_{\theta_t}}(s))] \\
% &= \mathbb{E}_{a \sim \pi_{\theta_t}(\cdot | s)}[\nabla_{\theta_t} \log \pi_{\theta_t}(a | s) A^{\pi_{\theta_t}}(s,a)]. \label{eq:aac}
% \end{align}
\begin{align}
% \nabla_{\theta_t} J(\theta_t) &= \mathbb{E}_{a \sim \pi_{\theta_t}(\cdot | s)}[\nabla_{\theta_t} \log \pi_{\theta_t}(a | s) (Q^{\pi_{\theta_t}}(s,a) - V^{\pi_{\theta_t}}(s))] \\
\nabla_{\theta_t} J(\theta_t) &= \mathbb{E}_{a \sim \pi_{\theta_t}(\cdot | s)}[\nabla_{\theta_t} \log \pi_{\theta_t}(a | s) A^{\pi_{\theta_t}}(s,a)], \\
A^{\pi_{\theta_t}}(s,a) &= Q^{\pi_{\theta_t}}(s,a) - V^{\pi_{\theta_t}}(s).
\label{eq:aac}
\end{align}
Here, the advantage function, $A^{\pi_{\theta_t}}(s,a)$, captures the extent to which a particular action choice from a given state offers greater value than that of the current policy.
% , executed from the same state.
Recently, actor-critic approaches have found great success in leveraging this advantage actor-critic formulation~\cite{Mnih2016AsynchronousMF}.

\subsection{COACH}
\label{sec:coach}
COACH is an actor-critic algorithm for HRL based on the insight that human feedback resembles the advantage function. Accordingly, the core update equation for COACH can be obtained by simply replacing the advantage function in Equation~\ref{eq:aac} with the observed human feedback:
\begin{align}
\label{eq:coach}
\nabla_{\theta_t} J(\theta_t) &= \mathbb{E}_{a \sim \pi_{\theta_t}(\cdot | s)}[\nabla_{\theta_t} \log \pi_{\theta_t}(a | s) f_t].
\end{align}
Recall that, since the human feedback $f_t \in \{-1, 0, 1\}$, it merely serves as a guide for the direction in which to follow the policy gradient. In addition, the COACH algorithm includes a human delay factor hyperparameter, $d$, which is the number of timesteps we believe captures the natural latency of a trainer observing behavior at a particular timestep and needing some small amount of reaction time before actually delivering his or her critique of the behavior. A proper setting of this trainer and domain-specific parameter is necessary to maintain a good alignment between observed feedback and state features. While the authors of COACH found 6 timesteps (0.2s) to be appropriate, we found 1 timestep (just over 1s) to be successful for our domain and trainers.

\subsection{Eligibility Traces}

While Equation~\ref{eq:coach} specifies a clear process by which to make successive updates to the parameters of an actor, it also yields a strong dependence on the observed human feedback. Clearly, it is impossible to receive feedback from a human trainer on every timestep and, in all likelihood, the feedback that is observed will be quite sparse, as the human trainer spends a certain amount of time observing the policy being executed by the agent before scoring the behavior. Crucially, COACH updates are not made based solely on the current policy gradient at each timestep, but rather based on an eligibility trace $e_\lambda$~\cite{Barto1983NeuronlikeAE}. The eligibility trace is an exponentially decaying accumulator of past policy gradients that allows for smoothing observed human feedback over a series of past transitions:
\begin{align}
e_\lambda &= \lambda e_\lambda + \nabla_{\theta_t} \log \pi_{\theta_t}(a_t | s_t).
\label{eq:coach_elig}
\end{align}
This mechanism helps deal with the challenge of sparse human feedback and allows an agent to reason about the quality of transitions where no explicit feedback was observed; in turn, a human trainer gains the ability to critique whole sequences of actions. 

To make COACH more amenable to the use of neural networks, we give up the idea of a single, continuous eligibility trace for all observed experiences in favor of an eligibility replay buffer. In Deep Q-Networks \cite{Mnih2015HumanlevelCT}, an experience replay buffer is a FIFO queue used for storing individual agent transitions so they may be resampled for multiple training updates. Consequently, the agent is able to glean more information from each individual experience while also enabling the efficiency that comes from batch learning. Deep COACH extends this idea using a replay buffer where the atomic elements stored are whole windows of experience, rather than individual transitions. A single window of experience is characterized by having a length (defined by the total number of constituent environment transitions) less than or equal to a window size hyperparameter, $L$, and by ending with a transition where the associated human feedback signal is non-zero; all other transitions within the window have no associated feedback. Accordingly, each time a human trainer elects to provide feedback, he or she completes an entire window of experience that is then stored in the buffer for subsequent training updates. For each uniformly sampled window from the buffer, Equation \ref{eq:coach_elig} is used to compute an eligibility trace over each window; the eligibility traces are then averaged over the entire minibatch and the mean eligibility trace is applied as a single update to the policy network parameters.

Finally, we draw attention to the fact that the use of a replay buffer creates a discrepancy between the policy being optimized at the current timestep and the policy (or policies) under which data sampled from the buffer was generated. To account for the inherent off-policy learning within the training updates, we leverage importance sampling ratios as outlined by \citet{Degris2012OffPolicyA}. Specifically, each transition stored within the windows of the eligibility replay buffer includes the probability with which the selected action was taken from the given state, under the policy at the time the experience occurred. Thus, when appending an experience that occurred at timestep $t^{'} < t$ to an eligibility trace for a sampled window, the update occurs as follows:
\begin{align}
e_\lambda &= \lambda e_\lambda + \frac{\pi_{\theta_t}(a_{t^{'}} | s_{t^{'}})}{\pi_{\theta_{t^{'}}}(a_{t^{'}} | s_{t^{'}})} \nabla_{\theta_t} \log \pi_{\theta_t}(a_{t^{'}} | s_{t^{'}}),
\label{eq:dcoach_elig}
\end{align}
where $\pi_{\theta_{t^{'}}}$ is the behavior policy \cite{Degris2012OffPolicyA}.

\subsection{Unsupervised Pre-training}

Given the competing desires for learning policies from little feedback and over high-dimensional observation spaces, we perform unsupervised pre-training of our policy network to reduce training time and initialize the policy network with good image features. More concretely, the policy is implemented as a convolutional neural network (CNN), where the first five layers initially serve as the encoder model within a convolutional autoencoder (CAE)~\cite{Hinton2006ReducingTD,Masci2011StackedCA}. A CAE is defined by a pair of functions $(f_{\theta_e}, g_{\theta_d})$ where $f_{\theta_e}$ is a neural network mapping a raw observation, $\mathbf{x}$, to some lower dimensional encoding and $g_{\theta_d}$ is a neural network that decodes the output encoding of $f_{\theta_e}$ to reconstruct the original observation, $\mathbf{x}$. Given a minibatch of $n$ samples, the model can be trained end-to-end by minimizing the reconstruction loss:
\begin{align}
\mathcal{L}(\mathbf{x}) &= \frac{1}{n} \sum\limits_{i=1}^n (g_{\theta_d}(f_{\theta_e}(x_i)) - x_i)^2.
\label{eq:cae_obj}
\end{align}

\begin{figure}
\centering
\includegraphics[width=\linewidth]{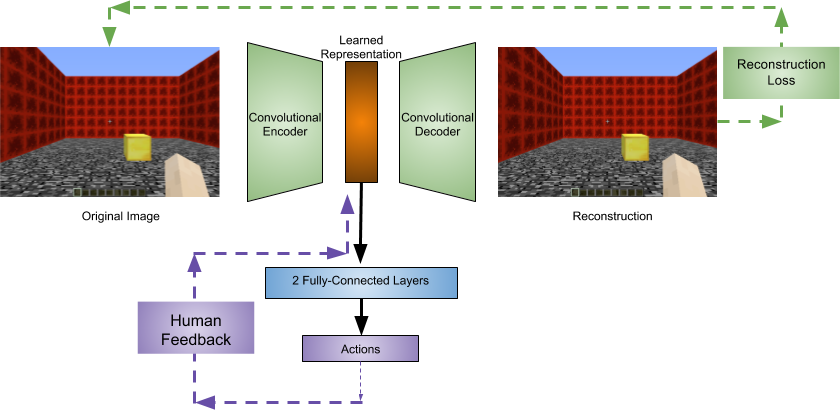}
\caption{Policy-network architecture and outline of training procedure. Dashed arrows indicate flow of training signals. Note that the training signal specified by human feedback only goes through the fully-connected layers of the policy network and does not affect the convolutional encoder parameters.}
\label{fig:training}
\end{figure}

In practice, the learned encoder model reduces the dimensionality of raw data while preserving relevant features of the original input that are critical to producing high-fidelity reconstructions. Accordingly, we treat the pre-trained encoder model as the front end of our policy network where the parameters of the encoder are treated as inputs to the algorithm and held fixed for the duration of learning (see Figure \ref{fig:training}). We note that the choice of leaving these parameters unperturbed by subsequent gradient updates based upon actual human feedback is more conducive to a lifelong learning setting and mitigates the risk of unrecoverable catastrophic forgetting \citep{French1999CatastrophicFI}; the generalized representation learned by the CAE can be used for training various tasks within a particular domain, only requiring a shift in the trainer's feedback signals and without having to reinitialize the algorithm.

\subsection{Entropy Regularization}

To prevent the agent from spuriously executing mixed sequences of actions (thereby making it difficult for the observing human trainer to accurately provide feedback), we greedily select the action with highest probability under the current policy at each timestep, instead of randomly sampling from the distribution. This action-selection policy is consistent with that of COACH \citep{MacGlashan2017InteractiveLF} and affords an opportunity to improve the responsiveness of Deep COACH to trainer feedback. While the theory of the off-policy learning update \citep{Degris2012OffPolicyA} demands that actions be sampled randomly, the resulting inability to interpret agent behavior and provide feedback represents a significant usability cost. Moreover, usage of the update outlined in Equation \ref{eq:dcoach_elig} still captures the spirit of how we would like the current policy to change in light of past experience.
% ; this kind of behavior policy could also be achieved under a softmax exploration policy with a high enough setting of the temperature parameter. 
% This method of action selection also affords a convenient opportunity to improve the responsiveness of Deep COACH to trainer feedback. While the theory of the off-policy learning update \citep{Degris2012OffPolicyA} demands that actions be sampled randomly, the resulting inability to interpret agent behavior and provide feedback represents a significant usability cost. Moreover, usage of the update outlined in Equation \ref{eq:dcoach_elig} still captures the spirit of how we would like the current policy to change in light of past experience.

A trainer who cannot observe the impact of his or her feedback in changing the agent's behavior will simply become frustrated; similarly, a trainer who must provide a large number of repetitive feedback cues to generate a sufficiently large number of gradient updates and induce a noticeable change in behavior will, most likely, give up on training before the task is learned. By virtue of optimizing with the policy gradient, consecutive updates shift probability mass towards actions that yield more positive feedback. However, inconsistencies and natural errors are inevitable when faced with human trainers \cite{Griffith2013PolicySI}. Additionally, depending on the window size parameter $L$, it is possible that a window of experience may contain behavior that the trainer did not mean to critique when supplying feedback. 

In expectation, we assume that the majority of feedback signals provided by the trainer are consistent. However, we would like to ensure that the policy is not prematurely biased towards misleading behavior while also making it relatively quick and simple for the trainer to alter the agent's current behavior. To accomplish this, we employ entropy regularization \cite{Williams1991FunctionOU} of the form $\beta \nabla_{\theta_t} H(\pi_{\theta_t}(\cdot | s_t))$ using a high regularization coefficient $\beta$ to maintain a high entropy policy. Accordingly, a policy that lies near uniform requires fewer successive updates to change the action with highest probability, thereby allowing for more immediate shifts in the agent's action-selection strategy and visible responsiveness to trainer feedback.

\begin{algorithm}[t]
\caption{Deep COACH}
\begin{small}
\begin{algorithmic}
\small
\STATE \textbf{Input:} Pre-trained convolutional encoder parameters $\theta_e$, human delay $d$, learning rate $\alpha$, eligibility decay $\lambda$, window size $L$, minibatch size $m$
\STATE Initialize eligibility replay buffer $\mathcal{E} \gets \emptyset$
\STATE Initialize window $\mathbf{w} \gets \{ \}$
\STATE Initialize first layers of $\pi_{\theta_0}$ with $\theta_e$ and freeze them 
\STATE Randomly initialize all remaining network parameters
\FOR{$t = 0$ to $\infty$}
	\STATE Observe current state $s_t$
	\STATE Execute action according to $a_t = \mbox{arg}\max\limits_a \pi_{\theta_t}(a | s_t)$
    \STATE Record $p_t \gets \pi_{\theta_t}(a_t | s_t)$
	\STATE Asynchronously collect human feedback $f_t$
    \STATE Append $\langle s_{t-d}, a_{t-d}, p_{t-d}, f_t \rangle$ to the end of $\mathbf{w}$
    \IF{$f_t \neq 0$}
    	\STATE Truncate $\mathbf{w}$ to the $L$ most recent entries and store in $\mathcal{E}$
%         \STATE Store $\mathbf{w}$ in $\mathcal{E}$
        \STATE $\mathbf{w} \gets \{ \}$
    \ENDIF
    \STATE Randomly sample a minibatch $W$ of $m$ windows from $\mathcal{E}$
    \STATE $\bar{e}_\lambda \gets 0$
    \FOR{$w \in W$}
    	\STATE Initialize eligibility trace $e_\lambda \gets 0$
        \STATE Retrieve final feedback signal $F$ from $w$
    	\FOR{$s, a, p, f \in w$}
        \STATE $e_\lambda \gets \lambda e_\lambda + \frac{\pi_{\theta_t}(a|s)}{p} \nabla_{\theta_t} \log \pi_{\theta_t}(a | s) $
        \ENDFOR
        \STATE $\bar{e}_\lambda \gets \bar{e}_\lambda + F e_\lambda$
    \ENDFOR
    \STATE $\bar{e}_\lambda \gets \frac{1}{m} \bar{e}_\lambda + \beta \nabla_{\theta_t} H(\pi_{\theta_t}(\cdot | s_t))$
    \STATE $\theta_{t+1} \gets \theta_t + \alpha \bar{e}_\lambda$
\ENDFOR
\end{algorithmic}
\end{small}
\label{alg:deepcoach}
\end{algorithm}

\subsection{Deep COACH}

The full Deep COACH algorithm is shown in Algorithm~\ref{alg:deepcoach} with an outline of the online training procedure in Figure~\ref{fig:training}. To maintain robustness and responsiveness, we restrict Deep COACH to relatively small policy network architectures with respect to both the number of layers and the number of hidden units per layer. Given that the policy network architecture begins with fixed, pre-trained layers of a convolutional encoder, this constraint on the number of parameters per layer solely applies to the remaining fully-connected layers in the network that map the encoded state representation to a probability distribution over the agent's action space. In all of our experiments, the portion of the policy network not trained after the unsupervised pre-training phase was represented by two fully-connected layers with no more than 30 hidden units. The convolutional encoder used in all experiments is identical to the architecture used in \cite{Mnih2015HumanlevelCT} except with only 256 units in the penultimate fully-connected layer and an output layer of 100 units.

\begin{figure}[t]
  \centering
  \includegraphics[width=0.6\linewidth]{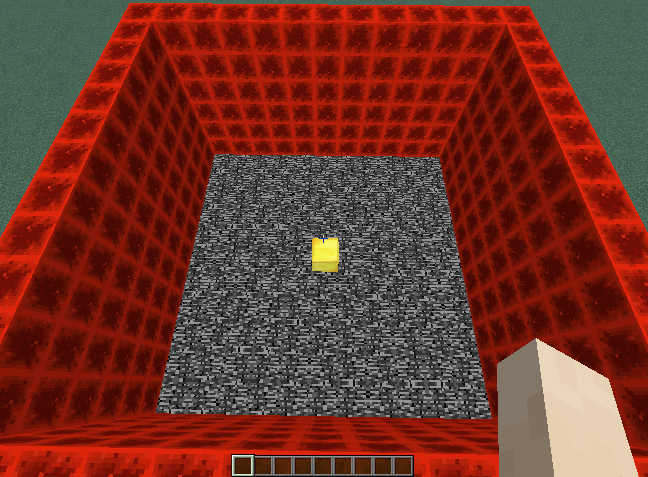}
  \caption{Top-down view of the Minecraft domain used in both evaluation tasks with a fixed gold block in the center.}
  \label{fig:minecraft_world}
\end{figure}

\begin{figure*}[t]
\begin{subfigure}{.3\linewidth}
  \centering
  \includegraphics[width=\linewidth]{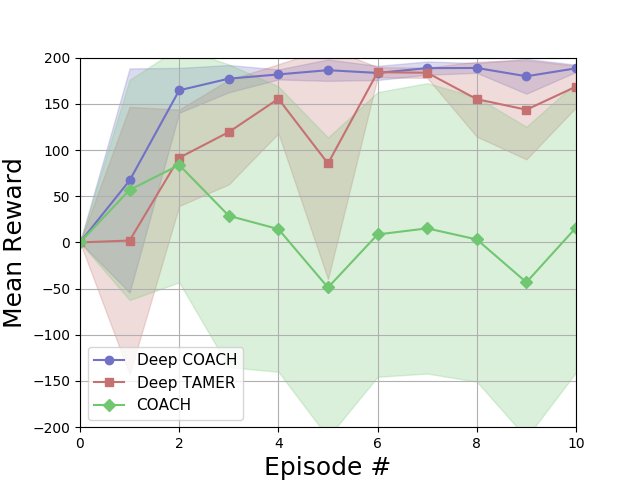}
  \caption{Mean episodic reward for the Goal-Navigation Task.}
  \label{fig:goal_nav_reward}
\end{subfigure}
\hfill 
\begin{subfigure}{.3\linewidth}
  \centering
  \includegraphics[width=\linewidth]{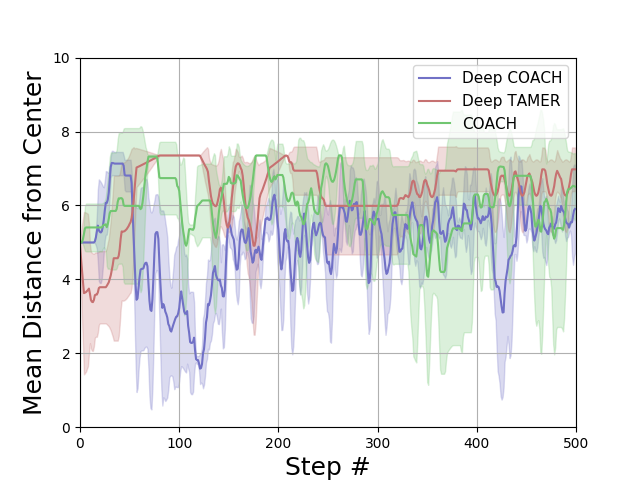}
  \caption{Mean Euclidean distance from center point in the Perimeter-Patrol Task.}
  \label{fig:patrol_dist}
\end{subfigure}
\hfill
\begin{subfigure}{.3\linewidth}
  \centering
  \includegraphics[width=\linewidth]{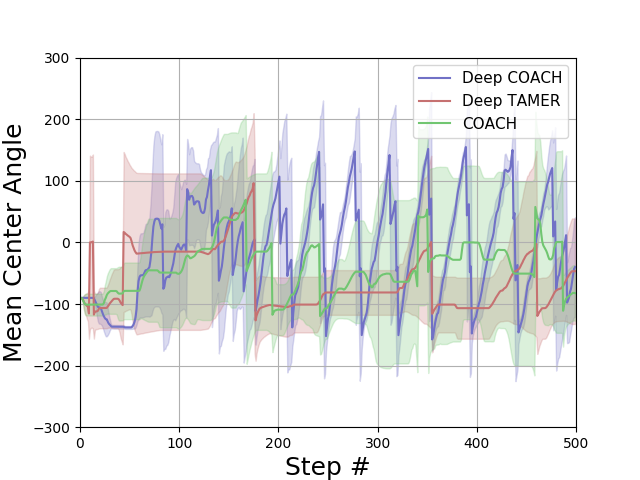}
  \caption{Mean angle (deg.) formed by agent's current position, start position, and grid center for the Perimeter-Patrol Task.}
  \label{fig:patrol_angle}
\end{subfigure}
\caption{Environment reward and task metrics collected during agent training.}
\end{figure*}

\section{Experiments}
To demonstrate the effectiveness of our Deep COACH as an improvement over the existing COACH algorithm, we present results on two illustrative tasks where human trainers (two authors of this paper) manage to elicit desired behavior from the learning agent after providing fewer than one hundred feedback cues. Both trainers performed similarly, and so we report results over their combined data. We leave the task of examining evaluative feedback training with uninitiated users to future work; the goal of this study, just as in other deep HRL papers \citep{Warnell2017DeepTAMER}, is to highlight the possible benefits of this HRL paradigm and not to argue that most users would be successful. In both tasks, the Deep COACH agent converges on satisfactory behavior within hundreds of timesteps corresponding to between 10 and 15 minutes of training time \footnote{We used a feature in Project Malm{\"o} to slow the internal speed of Minecraft thereby supplying ample trainer reaction time. Normally, each tick of the Minecraft clock is 50ms realtime; we internally slowed this to 250ms (4 actions per second).}.
% ; clearly, this result represents an improvement over the expected performance of any traditional deep reinforcement-learning agent learning from raw images for the same amount of time. 
Moreover, we implement and evaluate a Deep TAMER agent as outlined in \citet{Warnell2017DeepTAMER} and report its performance as well. \footnote{To mitigate potential biases, a several hour gap was maintained between each trial which took the form of a single warmup run prior to the reported run.} With the goal of presenting Deep COACH as a viable alternative for leveraging human feedback when learning from high-dimensional observations, we leave the investigation of robustness and success in the face of non-expert human trainers with no knowledge of the underlying training algorithm to future work.

All agents in our experiments were only ever provided with real-time feedback signals as specified by the human trainer. To obtain reportable performance metrics on the tasks, however, the environment (pictured in Figure \ref{fig:minecraft_world}) was given an underlying, task-specific reward structure that was recorded at each timestep but never made visible to the agent. 
% Furthermore, we note that a traditional, reward-maximizing reinforcement-learning agent learning our second task would experience difficulty in finding a solution by observing Markovian rewards \cite{Littman2017EnvironmentIndependentTS}.

In all experiments, a single CAE was used to initialize the policy network after being trained to convergence using Adam~\cite{Kingma2014AdamAM} (learning rate of $0.001$, minibatches of size 32) with standard reconstruction loss (Equation \ref{eq:cae_obj}) from a dataset of 10,000 images collected by executing a random policy in the environment. All observations were represented by $84 \times 84$ RGB images while all networks were implemented in Tensorflow~\cite{Abadi2016TensorFlowAS}. To optimize the remaining policy-network parameters, we employed the RMSProp optimizer \cite{Tieleman2012RMSProp} along with a human delay factor $d = 1$
% \footnote{While COACH found 6 timesteps (0.2s) to be appropriate, we found 1 timestep (just over 1s) to be successful for our domain and trainer.}
% \footnote{The internal speed of the Minecraft world was marginally slowed through Project Malm{\"o} such that this low delay setting left ample time for the trainers to react.}
, learning rate of $\alpha = 0.00025$, eligibility decay $\lambda = 0.35$, window size $L = 10$, minibatch size $m=16$, and entropy regularization coefficient $\beta=1.5$. As outlined in \citet{MacGlashan2017InteractiveLF}, our baseline COACH agent was represented by a linear function approximator on top of the pre-trained CAE features, using the same learning rate and eligibility decay as Deep COACH while optimized with stochastic gradient descent \cite{Rumelhart1988}. The Deep TAMER reward network was identical to the policy network of Deep COACH (with the exception of a final softmax activation). Using the same learning rate as Deep COACH, the network was optimized using the Adam \cite{Kingma2014AdamAM} optimizer with a buffer update interval of 10 and a uniform credit assignment interval of $[0.2, 2.0]$. All replay buffers were given no limits and all parameters were selected as a result of a prolonged but informal tuning. All agents were given the same discrete action space consisting of \texttt{forward}, \texttt{rotate left}, and \texttt{rotate right} actions. Upon selecting a single action, the agent would move continuously in the direction specified by the action until a new action selection was made on the next timestep. 
% \footnote{Even with the internal slowing of Minecraft, the agent still executes roughly 200-300 steps within a minute.}

\subsection{Goal Navigation}
In this task, the agent is randomly placed in a $10 \times 10$ grid facing a single gold block objective in the center of the room (as shown in Figure~\ref{fig:minecraft_world}). The agent must navigate from its start location to the gold block. The underlying environment reward structure for the task silently provides a reward of $+200$ to the agent for reaching the gold block while each step taken by the agent has a cost of $-1$. Each episode runs until the agent finds the goal or until the agent reaches a quota of 200 steps. Thus, an agent either completes an episode by reaching the goal and earning non-negative reward or fails to do so earning an exact reward of $-200$.

\subsection{Perimeter Patrol}

For this task, the agent is in the same Minecraft world as for the Goal-Navigation Task. The trainer's desired behavior is for the agent to continuously walk the perimeter of the room in a clockwise fashion. Due to the non-Markovian nature of the task \cite{bacchus96,Littman2017EnvironmentIndependentTS}, it is uninformative to report summaries of Markovian rewards; accordingly, we report two complementary statistics that, together, illustrate the success (or failure) of an agent in achieving the desired motion. Specifically, at each timestep, we record the Euclidan distance from the agent's current position to the center of the room as well as the angle (measured in degrees) formed by the agent's fixed start position and current position to the center of the room. Consequently, an agent successfully circling the perimeter of the room will maintain little variation in its distance from the center of the room while its angle to the center will follow a consistent, oscillatory pattern. Note that a traditional reinforcement-learning agent incentivized with, for example, positive rewards at each corner and negative rewards for crossing the middle of the room, would only learn to loiter at a single corner and continuously collect positive reward.

\begin{figure*}[t]
\centering
\null\hfill
\begin{subfigure}{.3\linewidth}
%   \centering
  \includegraphics[width=\linewidth]{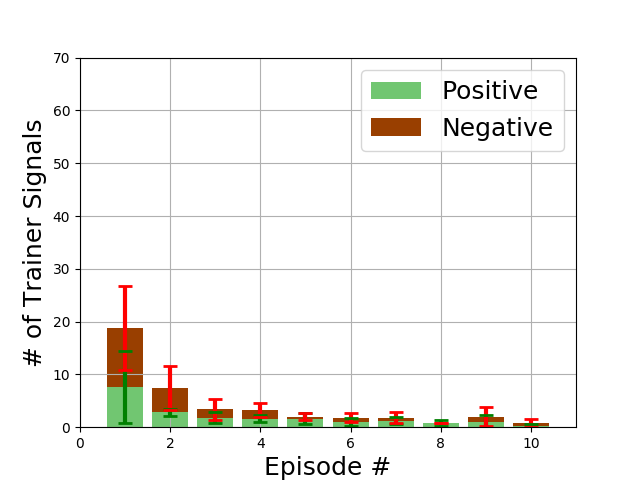}
  \caption{Deep COACH}
%   \label{fig:goal_nav_feed}
\end{subfigure}
\hfill 
\begin{subfigure}{.3\linewidth}
%   \centering
  \includegraphics[width=\linewidth]{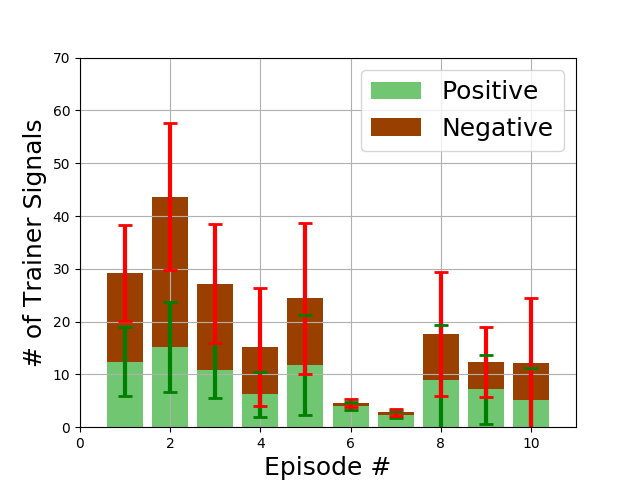}
    \caption{Deep TAMER}
    %  \label{fig:patrol_feed}
\end{subfigure}
\hfill\null
\caption{Breakdown of trainer feedback signals for the Goal-Navigation Task}
\label{fig:all_gw_feed}
\end{figure*}

\begin{figure*}[t]
\centering
\null\hfill
\begin{subfigure}{.3\linewidth}
%   \centering
  \includegraphics[width=\linewidth]{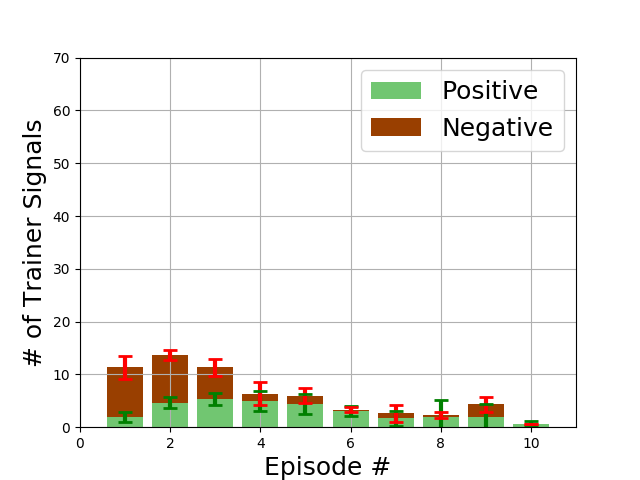}
  \caption{Deep COACH}
%   \label{fig:goal_nav_feed}
\end{subfigure}
\hfill 
\begin{subfigure}{.3\linewidth}
%   \centering
  \includegraphics[width=\linewidth]{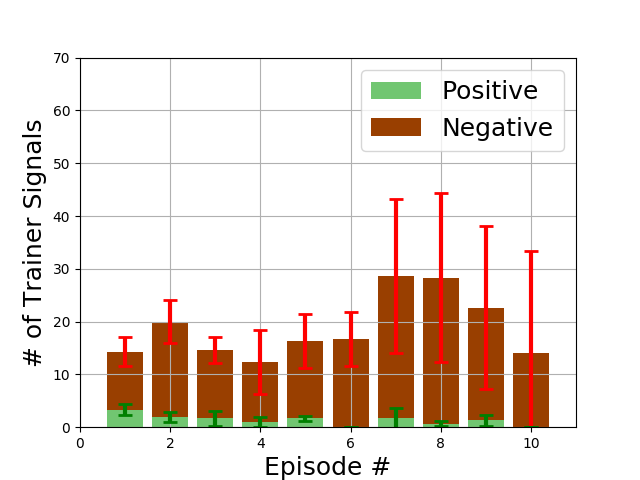}
    \caption{Deep TAMER}
    %  \label{fig:patrol_feed}
\end{subfigure}
\hfill\null
\caption{Breakdown of trainer feedback signals for the Perimeter-Patrol Task.}
\label{fig:all_patrol_feed}
\end{figure*}

\section{Results \& Discussion}

The mean episodic environment reward obtained by all algorithms over the course of five independent trials of the Goal-Navigation Task is shown in Figure~\ref{fig:goal_nav_reward}. Similarly, the mean step distances and angles (as specified in the previous section) obtained by each algorithm on three independent runs of the Perimeter-Patrol Task are shown in Figures ~\ref{fig:patrol_dist} and \ref{fig:patrol_angle}, respectively. Additionally, in Figures ~\ref{fig:all_gw_feed} and ~\ref{fig:all_patrol_feed}, we provide a full breakdown of human-trainer feedback for training Deep COACH and Deep TAMER respectively, separated into the positive and negative signals. All shading and error bars denote 95\% confidence intervals. For the Perimeter-Patrol Task, the 500 total steps taken by the agent in the domain are broken into chunks of 50 steps.

In the early episodes of the Goal-Navigation task, all agents experience a fairly high degree of variance in performance while accruing the experience and human feedback needed to isolate the desired task. After completing just a few episodes, however, both the Deep COACH and Deep TAMER agents manage to arrive at a policy that consistently reaches the target. 
% While Deep COACH is able to sustain this degree of performance, Deep TAMER demonstrates a certain degree of instability. 
Alternatively, the COACH agent, while capable of completing whole episodes of the task, suffers from a complete failure to generalize, resulting in episodes where the agent fails to reach the goal at all. Even with a small learning rate, we noticed that COACH could occasionally become ``locked'' into a policy with such low entropy that no amount of feedback was capable of shifting a sufficient amount of probability mass away from the single, repeating action; we suspect that this behavior is an artifact of mapping directly from a high-dimensional input to a low-dimensional output space. Throughout the course of learning, we see that Deep COACH trainer feedback (shown in Figure \ref{fig:all_gw_feed}) is policy-dependent in that it gradually decreases as the agent hones in on the desired behavior. Furthermore, the vast majority of trainer feedback consists of negative signals discouraging incorrect behavior at the start of learning whereas a relatively smaller number of positive signals are actually needed to guide the agent towards good behavior. While Deep TAMER feedback also decreases with time, we found that, in the early stages of learning, Deep TAMER required a substantial number of feedback cues before demonstrating any kind of shift in behavior.

At the start of learning for the Perimeter-Patrol Task, all agents experience difficulty with the presented task; the fluctuating, relatively constant distance from the center in the first few hundred steps suggests time spent by the agents either crossing the room or loitering at a corner. Moreover, the inconsistencies in the corresponding angles suggest no deliberate attempt at cyclic motion throughout the room. As more time passes, all agents come to maintain a fairly consistent distance from the center point. However, Deep COACH displays the corresponding oscillatory pattern in the angles of its trajectory, indicative of a patrolling motion through the room. (The rightward sloping sawtooth pattern is consistent with clockwise patrolling.) Note that the angles taken by Deep TAMER oscillate at a considerably lower frequency indicating a traversal of the perimeter with frequent stops. The trajectories produced by the COACH agent maintain little regularity and further highlight the challenge of learning the required policy through linear function-approximation techniques. 

Towards the end of learning, there are drops in the center distance maintained by the Deep COACH agent, indicating an erroneous crossing into or through the middle of the room. These were the result of minor catastrophic forgetting~\cite{French1999CatastrophicFI},  where a small set of updates from replay memory resulted in a temporary lapse in behavior. Fortunately, in almost all of these situations, the Deep COACH agent was able to correct itself after a few subsequent minibatch updates. In contrast, Deep TAMER encountered trials where forgetting would set in and become irreversible. 

As with the Goal-Navigation Task, Figure~\ref{fig:all_patrol_feed} captures a consistent decline in the total amount of feedback needed by the Deep COACH agent over the course of learning. Interestingly, in the Deep TAMER agent, we found that the presence of opposing rotation actions clashed with the overlapping credit assignment intervals of successive feedback signals, resulting in degenerate looping behavior between the two actions. Towards the end of learning, and prefaced by an uptick in feedback to correct occasional forgetting, we observe a near complete lack of human feedback as the Deep COACH agent displays the correct behavior and the trainer ceases to provide further positive reinforcement.

\section{Conclusion}

We present a novel algorithm for deep reinforcement learning of behavior from policy-dependent human feedback. We extended the existing COACH algorithm with a series of alterations that allow for scaling up to learning behavior from high-dimensional observations without incurring the large time and sample complexities typically associated with deep learning approaches: a modified replay memory buffer, the use of an autoencoder, and high entropy regularization. To demonstrate the efficacy of our approach, we presented results on two tasks grounded in the game world of Minecraft and show behaviors learned purely from trainer feedback signals within 15 minutes of interaction. Moreover, the desired behaviors were achieved with fewer than one hundred feedback signals from the live human trainers and surpass the performance of baseline approaches. We further explored the breakdown of human feedback over the course of learning with our algorithm to both verify that feedback gradually decreases over time as learning progresses and confirm that the agent does eventually converge to a policy that satisfies the human trainer.

One immediate direction of future work is to perform a thorough user study comparing the Deep COACH and Deep TAMER approaches on several tasks with non-expert trainers. Crucially, the comparison should extend beyond only task performance metrics and include breakdowns of trainer feedback that help quantify the ease of training and degree of trainer satisfaction under each algorithm. 

\bibliographystyle{icml2018}
\bibliography{references}

\end{document}